\documentclass[manuscript,screen]{acmart}

\AtBeginDocument{%
  }

\acmISBN{978-1-4503-XXXX-X/2018/06}

\usepackage[normalem]{ulem}  
\usepackage{multirow}
\usepackage{graphicx}
\usepackage{threeparttable}  
\usepackage{booktabs, multirow, threeparttable}
\usepackage{makecell} 
\usepackage{pgfplots}
\usepgfplotslibrary{fillbetween} 

\usepgfplotslibrary{groupplots}

\usepackage{pgfplotstable}

\usepackage{pgfplots}
\usepackage{xcolor}
\usepackage{subcaption}
\usepackage[abs]{overpic} 

\usepackage{tabularx}

\usetikzlibrary{intersections,fillbetween}
\usetikzlibrary{patterns}

\begin{document}

\title{Density-Guided Response Optimization: Community-Grounded Alignment via Implicit Acceptance Signals}

\author{Patrick Gerard}
\email{pgerard@isi.edu}
\affiliation{%
  \institution{Information Sciences Institute, University of Southern California}
  \city{Marina Del Rey}
  \state{California}
  \country{USA}
}

\author{Svitlana Volkova}
\affiliation{%
  \institution{Aptima Inc.}
  \city{Dayton}
  \state{Ohio}
  \country{USA}}
\email{svolkova@aptima.com}

\renewcommand{\shortauthors}{Gerard et al.}

\begin{abstract}

Language models deployed in online communities must adapt to norms that vary across social, cultural, and domain-specific contexts. Prior alignment approaches rely on explicit preference supervision or predefined principles, which are effective for well-resourced settings but exclude most online communities---particularly those without institutional backing, annotation infrastructure, or organized around sensitive topics---where preference elicitation is costly, ethically fraught, or culturally misaligned.

We observe that communities already express preferences implicitly through what content they accept, engage with, and allow to persist. We show that this acceptance behavior induces measurable geometric structure in representation space: accepted responses occupy coherent, high-density regions that reflect community-specific norms, while rejected content falls in sparser or misaligned areas. We operationalize this structure as an implicit preference signal for alignment and introduce \textit{density-guided response optimization} (DGRO), a method that aligns language models to community norms without requiring explicit preference labels.

Using labeled preference data, we demonstrate that local density recovers pairwise community judgments, indicating that geometric structure encodes meaningful preference signal. We then apply DGRO in annotation-scarce settings across diverse communities spanning platform, topic, and language. DGRO-aligned models consistently produce responses preferred by human annotators, domain experts, and model-based judges over supervised and prompt-based baselines. We position DGRO as a practical alignment alternative for communities where explicit preference supervision is unavailable or misaligned with situated practices, and discuss the implications and risks of learning from emergent acceptance behavior.

\end{abstract}





\maketitle

\definecolor{co-url}{HTML}{b8b8ff}
\definecolor{co-hashtag}{HTML}{ffa9a3}
\definecolor{co-retweet}{HTML}{ffac81}
\definecolor{co-fast-retweet}{HTML}{ace1af}
\definecolor{kNN_Embedding_Graph}{HTML}{afcafc}
\definecolor{tkNN_Embedding_Graph}{HTML}{d3d3d3}
\definecolor{Embedding_Averaging}{HTML}{0075F2}
\definecolor{kNN_Embedding_Graph}{HTML}{4CB5AE}

\definecolor{NoBridge}{HTML}{0075F2}        
\definecolor{Bridge}{HTML}{4CB5AE}

\section{Introduction}

Language models increasingly interact with online communities whose norms, values, and communicative conventions vary widely across social, cultural, and domain-specific contexts. What counts as an appropriate response depends not only on topic, but on situated expectations around tone, evidence, empathy, authority, and care. A question about weight loss, for example, calls for fundamentally different responses in a medical advice forum, a peer support community, or an academic discussion space—not because the underlying facts differ, but because the social meanings and potential harms of speech differ across contexts. Capturing these distinctions is essential not only for safe and effective deployment of language models, but also for broader questions of algorithmic governance: who defines acceptable behavior, whose values are encoded, and how those values are operationalized in deployed systems.

Existing approaches to language model alignment have largely addressed these questions through explicit preference supervision. Reinforcement Learning from Human Feedback (RLHF) and related methods rely on annotated preference data to guide model behavior \citep{christiano2017deep,ouyang2022training}, while Direct Preference Optimization (DPO) simplifies optimization but retains dependence on labeled comparisons \citep{rafailov2023direct}. Constitutional AI further reduces human annotation by introducing principle-based critiques \citep{bai2022constitutional}. While effective in settings where preferences can be clearly articulated and externally specified, these approaches presuppose that normative criteria are stable, consensual, and ethically straightforward to elicit. In practice, however, many online communities—particularly marginalized, informal, or sensitive ones—lack the institutional capacity, shared language, or ethical conditions required for explicit preference annotation. In such settings, asking external annotators to define “appropriate” behavior risks misrepresentation, cultural mismatch, or harm.

At the same time, community norms are not unexpressed. Online communities continuously enact and negotiate standards of appropriateness through moderation, participation, and collective attention. Content that aligns with community expectations is more likely to persist, receive engagement, and become part of ongoing discourse, while misaligned content is ignored, down-ranked, or removed. Importantly, these acceptance patterns are shaped not only by individual preferences, but also by power, platform affordances, and governance structures within the community. As such, behavioral acceptance should not be treated as normative endorsement or consent. Rather, it constitutes a descriptive signal of how norms are operationalized in practice, reflecting the values of those who are most able or willing to participate.

Building on prior work on implicit behavioral signals in recommender systems and information retrieval \citep{hu2008collaborative,joachims2007evaluating}, we study whether these naturally occurring acceptance patterns give rise to recoverable structure in representation space. We observe that responses accepted by a community are not randomly distributed; instead, they tend to cluster in coherent, high-density regions of embedding space, which we refer to as a community’s \emph{acceptance manifold}. This structure captures what a community treats as permissible or contextually normal, as enacted through collective behavior rather than prescribed by external rules. We emphasize that this manifold reflects descriptive regularities in community practice, not an ethical claim about which norms ought to be learned or deployed.

We operationalize this observation through \textbf{Density-Guided Response Optimization (DGRO)}, a method that uses local density in a community’s embedding space as an implicit preference signal for alignment. DGRO does not assume that community norms are universally desirable or stable; instead, it provides a mechanism for studying and modeling how norms manifest in behavior when explicit preference supervision is unavailable or inappropriate. We first validate the underlying manifold hypothesis on labeled preference data, showing that local density correlates monotonically with observed human judgments. We then demonstrate that this signal can substitute for explicit preference annotations within standard preference optimization objectives. Finally, we apply DGRO in annotation-scarce settings across diverse communities, including eating disorder support spaces and Russian-language conflict documentation forums, and evaluate whether aligned models produce responses judged as more contextually appropriate and authentic.

This work makes three contributions. First, we provide empirical evidence that community acceptance behavior induces structured, locally coherent geometry in representation space that encodes recoverable preference signal. Second, we introduce DGRO as a practical, annotation-free mechanism for leveraging this structure in preference-based alignment. Third, we analyze the ethical implications and limitations of learning from acceptance behavior, including risks of bias amplification, exclusion, and manipulation, and situate DGRO as a descriptive alignment tool whose deployment requires careful governance and oversight.

\section{Related Work}

\vspace{2pt} \noindent
\textbf{Alignment from Explicit Preferences}
Most modern alignment methods assume access to explicit human preference supervision.
Reinforcement Learning from Human Feedback (RLHF) learns a reward model from annotated
pairwise comparisons and optimizes a policy via reinforcement learning
\citep{christiano2017deep,ouyang2022training}. While effective, this paradigm requires
large volumes of carefully curated preference data and a multi-stage training pipeline.
Direct Preference Optimization (DPO) simplifies optimization by removing the reward model
and reinforcement learning stage, but still fundamentally relies on explicit preference
labels \citep{rafailov2023direct}. Constitutional AI further reduces human annotation by
substituting AI-generated critiques guided by predefined principles \citep{bai2022constitutional},
yet this shifts the burden to principle specification and presumes that normative criteria
can be articulated a priori. Across these approaches, alignment is framed as supervised
learning from observable preferences, limiting applicability in settings where preferences
are implicit, emergent, or difficult to elicit.

\vspace{2pt} \noindent
\textbf{Community Norms and Domain-Specific NLP}
A growing body of NLP research emphasizes the importance of cultural, social, and
community-specific norms, particularly in low-resource or marginalized contexts
\citep{bird2020decolonising,liu2025culturally}. Domain adaptation and specialization
techniques such as BioBERT and LegalBERT demonstrate the value of tailoring models to
specific domains, but typically require substantial labeled data
\citep{lee2020biobert,chalkidis2020legal,ben2010theory}. Ethical NLP work further argues
for embedding social values and community perspectives into model design
\citep{hutchinson2020social,leidner2017ethical,10.1145/3442188.3445922}, yet little work has explored how such norms can be learned operationally from naturally occurring community behavior. Our approach contributes a concrete mechanism for grounding alignment in community norms by inferring them directly from patterns of acceptance, without requiring explicit annotation or predefined value schemas.

Beyond linguistic variation, work in social computing and HCI highlights that community
norms are not merely emergent patterns of language use, but are actively shaped through
moderation practices, governance structures, and participation asymmetries
\citep{ostrom1990governing,jhaver2019does,matias2019preventing}. These dynamics raise
questions of legitimacy and representation: whose behavior contributes to observable
norms, and whose voices are systematically excluded. While prior NLP work has emphasized
the importance of respecting community norms, relatively little research has explored
how such norms can be operationally inferred from naturally occurring community behavior
without relying on explicit annotation or predefined value schemas.


\vspace{2pt} \noindent
\textbf{Implicit Behavioral Signals and (the Limits of) Revealed Preference}
A large body of work has explored learning from implicit behavioral signals, such
as clicks, dwell time, and interaction patterns, particularly in recommender
systems and information retrieval \citep{joachims2007evaluating,hu2008collaborative}.
These signals are attractive because they are abundant and naturally occurring, but
they are also indirect: they reflect behavior mediated by platform affordances,
incentives, and power rather than explicit judgments of quality or appropriateness.
Prior work has shown that optimizing directly for engagement can distort model behavior,
amplifying polarized, sensational, or emotionally charged content
\citep{barbera2020social}.

A long line of critique cautions against equating observed behavior with normative
endorsement or consent, particularly in platform-mediated environments
\citep{hands2014paul,matias2019preventing}. In this work, we therefore treat acceptance
signals as descriptive evidence of how norms are enacted in practice rather than as
ethically authoritative preferences. Our goal is not to maximize engagement or infer
individual utilities, but to recover community-level regularities in what is treated as
acceptable within specific contexts.

\vspace{2pt} \noindent
\textbf{Density and Geometry in Representation Space}
Density estimation has a long history in statistics, with classical approaches such as
kernel density estimation and Gaussian mixture models providing flexible non-parametric
tools \citep{parzen1962estimation}. Recent advances in neural density
estimation enable scalable likelihood modeling in high-dimensional spaces, including
autoregressive models and normalizing flows
\citep{rezende2015variational,dinh2016density,papamakarios2017masked}. Separately, work on
representation geometry in NLP has shown that linguistic representations occupy structured,
low-dimensional manifolds in embedding space \citep{arora2018linear,li2020sentence}. However, these techniques have primarily been used for generative modeling or representation analysis, rather than for norm inference or alignment. Building on these works, our approach interprets local density in embedding space as a community-conditioned acceptance signal, using geometric structure as supervision for alignment without explicit preference labels.


\section{Method}
\label{sec:method}

Our goal is to extract an alignment signal from naturally occurring community behavior without relying on explicit preference annotations. We build on the observation---well established in both social computing and recommender
systems---that communities already express preferences implicitly through what content they accept, engage with, and allow to persist. We show that repeated community acceptance induces measurable structure in representation space, and that this structure can be operationalized as a preference-aligned signal for language model alignment.

\begin{figure}[h]
\centering
\includegraphics[width=0.55\linewidth]{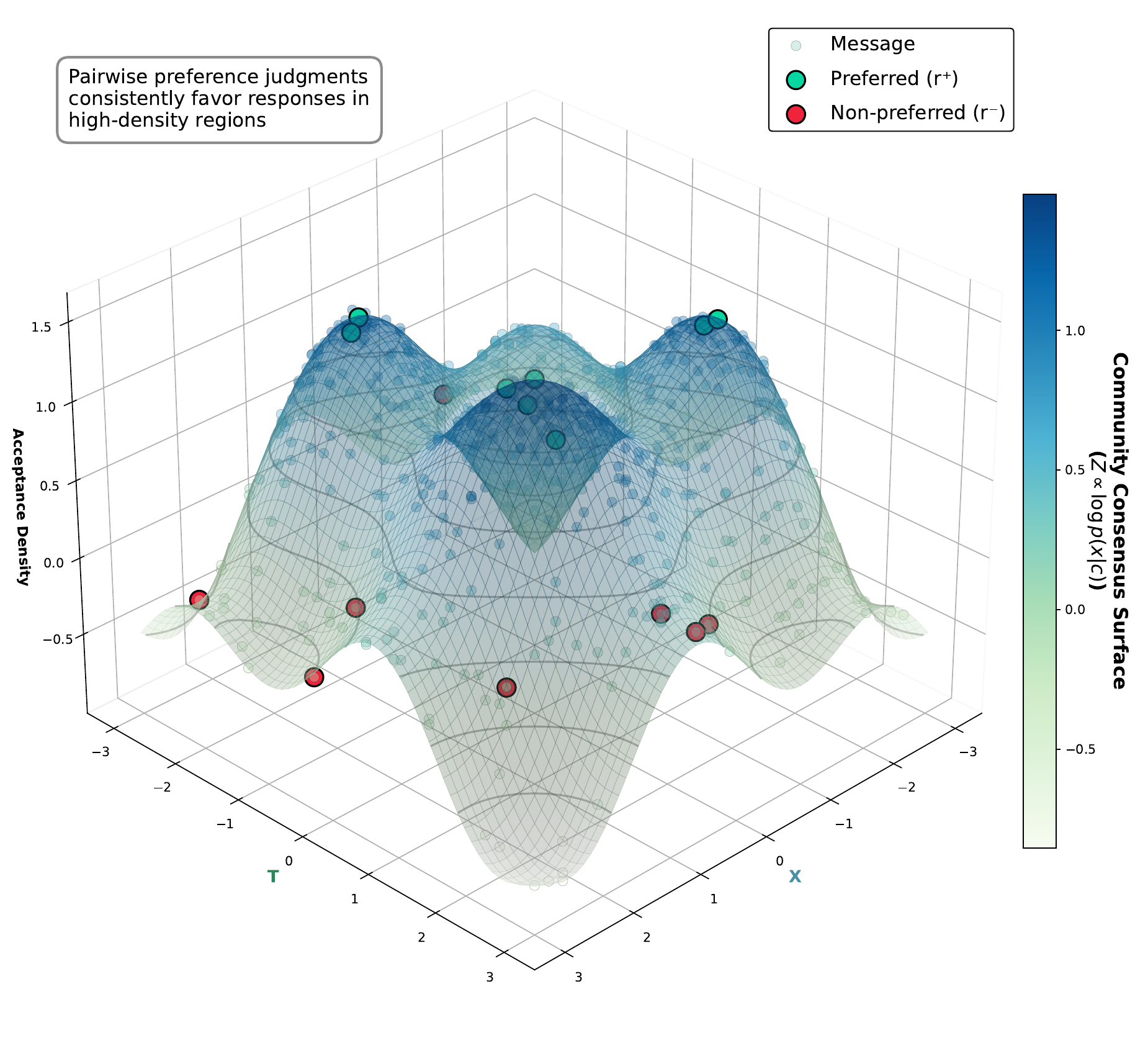}
\caption{\textbf{Conceptual Representation of the Community Consensus Surface.}
The Z-axis represents a normative log-density, reflecting the implicit filtering of
responses by community standards through moderation and collective feedback
\citep{lampe2004slash,chandrasekharan2017you}.
High-density regions correspond to a coherent, low-dimensional manifold of accepted
responses in representation space
\citep{arora2018linear,li2020sentence}.
The separation between preferred ($r^+$) and non-preferred ($r^-$) responses across
this surface reflects an \emph{acceptance--preference correspondence}, motivating
preference learning and alignment without explicit annotation
\citep{christiano2017deep,ouyang2022training,rafailov2023direct}.
}
\label{fig:consensus_surface}
\end{figure}

\subsection{Conceptualization: Community Acceptance as a Manifold}

Community norms are not imposed instantaneously; they emerge gradually through repeated interaction. Over time, online communities continuously filter participation through moderation, feedback, and collective attention. Responses that align with shared expectations are more likely to persist, receive engagement, and be incorporated into ongoing discourse. Responses that violate these expectations are disproportionately ignored, down-ranked, or removed.

This repeated process of selection acts as a form of implicit norm formation and expression. As similar responses are consistently accepted across comparable contexts, they accumulate and reinforce one another, giving rise to behavioral and linguistic regularities at the community level. For intuition, consider responses as points scattered across a 3D landscape, where elevation represents community acceptance density (Figure \ref{fig:consensus_surface}). Accepted responses---those that persist, receive engagement, or align with community norms---cluster in peaks of high density (high elevation), forming a coherent acceptance manifold. In contrast, rejected or misaligned responses lie in sparse, low-density regions at lower elevation, farther from the community’s normative core. This geometric separation mirrors the acceptance–preference correspondence illustrated in Figure~\ref{fig:consensus_surface}, where preferred ($r^+$) and non-preferred ($r^-$) responses occupy distinct regions of the surface. Prior work shows that such endogenous filtering dynamics produce durable patterns in language use, interaction style, and participation structure within communities
\citep{danescu2013no, horne2017identifying, chandrasekharan2018internet}.


We formalize this phenomenon geometrically, drawing on representation geometry~\citep{mikolov2013distributed, arora2018linear} and density-based clustering~\citep{ester1996density}, which show that linguistic and semantic structures occupy low-dimensional manifolds in embedding space. For a community $c$, we define an
\emph{acceptance manifold} $\mathcal{M}_c$ as the region of representation space occupied by responses that the community accepts as appropriate or authentic. Note that acceptance here is not a binary property of individual messages (e.g., receiving upvotes or avoiding removal), but an aggregate notion that emerges over time from patterns of participation and persistence within the community.
Let $E(r)$ denote the embedding of a response $r$. We model community acceptance
as a density over representations,
\[
p(r \mid c) = p(E(r) \mid c),
\]
where higher density indicates stronger conformity with community norms.
This view is consistent with distributional perspectives on language, in which
semantic and pragmatic regularities correspond to geometric structure in
embedding space \citep{mikolov2013distributed, ethayarajh2019contextual}.
Here, however, geometry reflects not only semantic similarity, but also
normative compatibility with a specific community.

The gradient of the log-density,
\[
\nabla_{E(r)} \log p(E(r) \mid c),
\]
defines a continuous direction of increasing alignment with community norms.
Unlike discrete preference labels, this signal is smooth, shared across
responses, and derived directly from observed behavior.

This framing induces an \emph{acceptance–preference correspondence} assumption:
responses that are repeatedly accepted by a community are more likely to align
with that community’s preferences. Formally,
\[
\arg\max_r p(r \mid c)
\;\propto\;
\arg\max_r \mathbb{E}[\text{preference}(r \mid c)].
\]

This assumption parallels foundational results in revealed preference theory
and implicit feedback learning, where aggregate behavioral signals—despite being
noisy at the individual level—can be used to \emph{empirically derive stable
community-level preferences and norms}
\citep{hands2014paul, houthakker1950revealed, hu2008collaborative, joachims2017unbiased}.
Here, we treat acceptance behavior as a revealed signal of \emph{collective consensus}: the norms that emerge from repeated, distributed decisions about what content is permitted, engaged with, persists within a community.

\subsection{Problem Formulation}

Let $\mathcal{D}_c = \{r_i\}_{i=1}^N$ denote a corpus of responses that have been
accepted by a community $c$ through moderation, engagement, or sustained
participation. We embed each response as $x_i = E(r_i)$ and interpret their
distribution in representation space as an empirical record of the community’s
acceptance behavior.

Our goal is to use it to \emph{derive an implicit preference signal}. Specifically, we view local acceptance density as inducing a partial ordering over candidate responses: responses that lie in higher-density regions of the acceptance manifold are more consistent with community norms than those in low-density regions.

In standard alignment pipelines such as RLHF \citep{christiano2017deep} or Direct Preference Optimization (DPO) \citep{rafailov2023direct}, learning is driven by explicit pairwise preference annotations. In contrast, we replace this supervision with a density-derived preference signal. For a given context, candidate responses can be ranked according to their relative \textit{acceptance density},
\[
p(E(r) \mid c),
\]
which serves as a proxy for community preference in the absence of human-labeled
comparisons.

We refer to this approach as \textbf{Density-Guided Response Optimization (DGRO)}. DGRO uses acceptance density to construct implicit preferred and dispreferred response pairs, enabling standard preference-based objectives such as DPO to be applied in annotation-scarce settings. This formulation aligns with prior work showing that geometric and distributional structure can substitute for direct supervision in low-resource regimes \citep{gretton2012kernel, arora2016latent}. 


\subsection{Operationalizing Acceptance Density}
\textit{Acceptance density} is a conceptual object defined over representation space. A key design choice is whether to estimate this density globally across all community content or locally conditioned on context. A global estimate implicitly assumes that community norms are uniform across topics and intents—a strong assumption that we later show obscures preference signal. We therefore adopt a \emph{local} density estimation strategy inspired by
neighborhood-based semantic modeling and local distributional structure
\citep{le2014distributed, huang2012improving}, while treating global density
estimation as a baseline.

Given a query context $h$ (e.g., a conversation history or post topic) with
embedding $E(h)$, we define a context-conditioned reference set
\[
\mathcal{B}(h) = \text{kNN}(h; \{E(h_i)\}_{i=1}^N),
\]
consisting of the $k$ nearest contexts. Let
$\{x_j\}_{j \in \mathcal{B}(h)}$ denote the embeddings of the corresponding
accepted responses.

We estimate acceptance density using a kernel density estimator,
\[
\log p(x \mid h, c) \propto
\log \frac{1}{|\mathcal{B}(h)|} \sum_{j \in \mathcal{mathcal {V}B}(h)} K_\sigma(x, x_j),
\]
where $K_\sigma$ is an RBF kernel with bandwidth set via the median heuristic.
This gives us a context-sensitive estimate: responses are scored relative to what
the community accepts in similar situations, rather than against an aggregated
global pool.

If acceptance density reflects community preference structure, it should both correlate with labeled human preference behavior in supervised settings and serve as a practical substitute for explicit preference annotations when used to train alignment objectives such
as DPO or RLHF. We evaluate both implications empirically in
Section~\ref{sec:validation} before deploying DGRO in annotation-scarce domains.

\section{Experimental Setup}
Our experiments are structured to answer three progressively stronger questions. First, we validate the \emph{manifold hypothesis}: whether community preference signals exhibit local geometric structure in representation space.
Next, we test whether \textit{acceptance density} can \emph{functionally replace} explicit human preference labels inside a standard optimization objective.
Finally, we evaluate whether this signal can be used to align language models in real-world communities where preference annotations are unavailable.

\subsection{Validating the Manifold Hypothesis}
\label{sec:manifold-validation}

First, we seek to validate the core
premise of our approach: that preference signal exhibits \emph{local geometric structure} in representation space. We use the Stanford Human Preferences (SHP) benchmark~\citep{ethayarajh2022understanding}, which provides pairwise preference judgments across Reddit communities as well as an external quality signal measuring the strength of human agreement.


\vspace{2pt} \noindent
\textbf{Communities and Data.}
We select five subreddits with clearly distinct moderation regimes and community norms: \textit{changemyview}, \textit{askculinary},
\textit{askhistorians}, \textit{legaladvice}, and \textit{explainlikeimfive}; these communities spanning different domains, interaction styles, and standards for acceptable responses. These communities differ substantially in how responses are evaluated, filtered, and endorsed, providing a controlled setting to test whether preference structure is shared across heterogeneous norms rather than driven by idiosyncrasies of a single community. Additional details about each community are provided in Appendix Table~\ref{tab:shp_subreddits}. Each example consists of a conversation history (prompt), a preferred response
and a non-preferred response as determined by community member voting, along with
metadata including the normalized ratio of upvotes between responses, which
captures preference strength.


\vspace{2pt} \noindent
\textbf{Testing the Manifold Hypothesis.} We ask whether responses preferred by a community tend to occupy higher-density
regions of representation space than non-preferred responses, when density is
estimated using only unlabeled data.
To test this, we first embed all responses from the training split, treating them
as an unlabeled reference pool that includes both preferred and non-preferred
responses. We use a fixed sentence encoder to obtain representations, enabling
density estimation over the resulting embedding space.\footnote{
We use the \texttt{sentence-transformers/all-mpnet-base-v2} encoder
(\url{https://huggingface.co/sentence-transformers/all-mpnet-base-v2}), a widely
adopted semantic model that provides stable neighborhood structure across
domains. Results are robust to alternative encoders; see
Appendix~\ref{app:embeddings}.
} Preference information is not used at this stage, and training and test splits are kept strictly disjoint. For each prompt in the test set, candidate responses are embedded and ranked according to their acceptance density under the community distribution. We then evaluate whether the response with higher estimated density corresponds to the community-preferred response.

\vspace{2pt} \noindent
\textbf{Evaluation Protocol.} For each test pair $(h, r_+, r_-)$, we compute a margin given by the difference in estimated acceptance density between $r_+$ and $r_-$. We report pairwise accuracy, $\mathbb{P}[\text{margin} > 0]$, as the primary metric. SHP provides the ratio of upvotes between responses normalized as an independent measure of community agreement strength. If preference signal is encoded in local geometry, our density-based margins should align with human preferences and improve as community agreement increases.


\vspace{2pt} \noindent
\textbf{Baseline Methods.} 
Our model, which we call acceptance density, estimates density conditioned on the $k=150$ nearest histories in embedding space; performance is robust to $k$ and we report ablations in Appendix~\ref{app:k-robustness}.

We compare against the following baselines. (1) Random assigns random margins as a sanity check. (2) $k$-Nearest Neighbors (kNN) retrieves the $k=150$ most similar training histories and predicts the majority preference label, testing whether neighborhood selection alone provides signal. (3) Global acceptance density estimates acceptance density using a fixed random subset ($|G|=1000$) of training responses, testing whether density modeling without locality recovers preference structure. Finally, we report results for the (4) original supervised SHP reward model~\footnote{\url{https://huggingface.co/stanfordnlp/SteamSHP-flan-t5-xl}}. This model serves as an upper-bound reference, illustrating how closely density-based methods trained on unlabeled data approximate preference signals learned from large-scale human annotations.

\subsection{Acceptance Density as a Preference Proxy}
\label{sec:preference-proxy-validation}

Building on the validation in the previous section, we next test whether
acceptance density can replace human-labeled comparisons within a standard
preference optimization pipeline, and whether doing so induces preference
behavior aligned with community judgments.

To test this, we instantiate a density-based variant of Direct Preference
Optimization (DPO) that uses acceptance density to construct implicit preference
supervision. We follow the same procedure for estimating acceptance density
described in the previous section, treating the training split as an unlabeled
reference pool and never using ground-truth preference labels during training.
Density-derived rankings are used to form implicit preferred and dispreferred
response pairs, which are then used to train a policy model with the standard
DPO objective.

Unless otherwise specified, all main results initialize from a pre-trained
Pythia-2.8B language model. This choice mirrors the experimental setup used in prior DPO work~\citep{rafailov2023direct}, which uses Pythia-2.8B~\cite{biderman2023pythia} as a primary reference architecture for
preference optimization; we do this for direct comparability and to isolate the effects
of the preference signal rather than architectural differences. Evaluation is performed on a held-out test split.

To assess robustness, we additionally repeat this procedure across multiple
model architectures and parameter scales. These results show consistent
trends, and we report deviations from the Pythia-2.8B baseline in
Appendix~\ref{app:model-robustness}.



\vspace{2pt} \noindent
\textbf{Evaluation protocol.} Evaluation is performed against \emph{held-out ground-truth human preferences}. We assess alignment using length-normalized preference accuracy, defined as the fraction of held-out SHP pairs for which the model assigns higher average log-probability per response token to the human-preferred answer. Log-probabilities are computed over response tokens only, conditioned on the shared prompt, ensuring that differences in response length do not confound the comparison. This evaluation directly tests whether optimization driven solely by acceptance density induces models to prefer the same responses that human annotators judge as better, which is the central objective of preference-based alignment. We report this metric for both supervised DPO (trained on true human preference pairs) and acceptance density-guided DPO under identical architectures, prompts, and evaluation conditions. This isolates a fundamental question: whether acceptance density behaves \emph{like a preference signal} when used as the sole source of supervision inside a standard alignment objective. Demonstrating competitive performance under this constraint establishes acceptance density as a viable substitute for explicit preference labels, justifying its use in annotation-scarce domains for alignment purposes.

\subsection{Application to Annotation-Scarce Communities}
\label{sec:application-to-communities}
Following the validation experiments above, we apply density-guided response
optimization (DGRO) in real-world communities where explicit preference
annotations are unavailable, and evaluate its effectiveness for aligning
language models in practice. In these settings, acceptance density is used to construct implicit preference
supervision. Using unlabeled community data, we estimate acceptance density as
described in Section~\ref{sec:manifold-validation} and use it to form preferred
and dispreferred response pairs. These density-derived pairs are then used to
train policy models with a standard DPO objective. No explicit pairwise
preference annotations are used at any stage.

\vspace{2pt} \noindent
\textbf{Communities and data.} We evaluate DGRO in settings where general-purpose models fail to capture domain-specific norms, and where standard preference annotation methods pose significant ethical risks.

Our primary evaluation focuses on eating disorder support communities across three platforms (Reddit, Twitter, and specialized forums). These communities exhibit highly sensitive, context-dependent communication norms distinct from general instruction-following behavior. Prior work indicates that off-the-shelf LMs often generate content that members find inauthentic or harmful~\citep{ziems2023normbank, vidgen2020directions, he2024community}. To address the ethical challenges of working in this domain, our data curation was conducted in collaboration with clinical domain experts and medical professionals as part of a broader study on online community formation (with IRB approval). Using expert-verified implicit signals avoids the ethical pitfalls of explicit annotation, including consent issues and potential re-traumatization.


To validate cross-lingual and political discourse generalization, we extend our evaluation to conflict documentation communities on VKontakte (VK), a Russian-language platform structurally comparable to Facebook~\citep{baran2015facebook}. These communities focus on the aggregation and discussion of ongoing conflict documentation, exhibiting norms distinct from both Western platforms and general Russian-language corpora. Current multilingual models, typically trained on broad web corpora, lack exposure to these specific discourse conventions. Using these data, we test DGRO's ability to adapt to distinct sociopolitical dialects where standard models often produce responses that appear foreign to the community's authentic communication patterns.

\begin{table}[t]
\centering
\caption{Communities and data sources used in DGRO evaluation.
Validation communities provide explicit preference supervision, while application
communities lack pairwise labels and rely on behavioral acceptance signals.}
\resizebox{0.64\columnwidth}{!}{%
\small
\setlength{\tabcolsep}{6pt}
\renewcommand{\arraystretch}{1.15}
\begin{tabular}{llcl}
\toprule
\textbf{Community} & \textbf{Platform} & \textbf{Scale} & \textbf{Acceptance Signal} \\
\midrule
Q\&A & Reddit (SHP) & 10K--50K pairs & Pairwise human preferences \\
\midrule
Eating Disorder Support & Twitter & $\sim$43K posts & Replies, retweets \\
Eating Disorder Support & Reddit & $\sim$9.2M posts & Upvotes, comment depth \\
Eating Disorder Support & Forums & $\sim$1.6M posts & Replies, thread continuation \\
Conflict Documentation & VK & $\sim$8.34M posts & Likes, reposts \\
\bottomrule
\end{tabular}
}
\label{tab:data_summary}
\end{table}

\vspace{2pt} \noindent
\textbf{Evaluation protocol.} 
The goal of this evaluation is to assess whether density-guided response optimization produces outputs that are judged as more appropriate and authentic within communities where explicit preference annotations are unavailable. As established in earlier sections, this analysis rests on two validated prerequisites: first, that acceptance density reliably recovers human pairwise preferences when such labels are available (Section~\ref{sec:manifold-validation}); and second, that density-guided optimization induces model behavior aligned with those same human judgments on held-out data (Section~\ref{sec:preference-proxy-validation}). Having validated both the preference signal and its effect on model behavior, we now evaluate aligned models in annotation-scarce domains.

Because these domains lack large-scale preference annotations, evaluation must rely on indirect judgments. We therefore anchor LLM-based evaluation in human expert assessment, first conducting expert evaluation on a stratified subset of 200 held-out examples (50 per domain), with three domain experts per community. This analysis verifies that aggregate LLM judgments track expert assessments along the same criteria. Following established practices in alignment research \cite{zheng2023judging, dubois2024length}, we then use LLM-as-judge comparisons along two criteria---relevance (contextual appropriateness to the prompt and community norms) and authenticity (consistency with the community’s characteristic tone, framing, and interactional style)---as a scaling mechanism for this previously validated human preference structure, rather than as an independent source of normative authority. Evaluation is performed in a head-to-head setting, where judges compare a model-generated response against an actual response drawn from the target community for the same context, using examples held out from all training stages. We use three frontier language models as judges: GPT-5-nano, Claude-4.5-Haiku, and Gemini-2.5-Flash.\footnote{
GPT-5-nano (\url{https://platform.openai.com/docs/models}),
Claude-4.5-Haiku (\url{https://www.anthropic.com/claude}),
and Gemini-2.5-Flash (\url{https://ai.google.dev/gemini-api/docs/models})
} Each model is queried three times with randomized response order to control for positional bias, yielding nine judgments per comparison.

\vspace{2pt} \noindent
\textbf{Baselines and model variants.} We compare DGRO against three baselines: (1) an off-the-shelf instruction-tuned model (Base), (2) supervised fine-tuning on community text (SFT), and (3) in-context learning with community exemplars (ICL). To isolate density-guided optimization from supervised pre-training effects, we conduct ablations controlling for training compute.

All comparisons use identical architectures, decoding parameters, and context construction. As explored in prior sections and further examined in Appendix~\ref{app:model-robustness}, variation across model architectures and scales appears limited for preference alignment under density-guided DPO. As such, we fix the base model to Pythia-2.8B in this section in order to focus on the behavioral and normative effects of the alignment procedure itself, rather than introducing additional variation from differences in model capacity or representation.

\section{Results}
\label{sec:validation}

\begin{table*}[th]

\centering
\smaller
\setlength{\tabcolsep}{6pt}
\caption{
Pairwise accuracy across communities for unsupervised and supervised methods.
Accuracy is reported as mean $\pm$ bootstrap half-width,
$\delta = \tfrac{1}{2}(\text{hi}-\text{lo})$, computed independently per subreddit.
Supervised Model (RM) denotes the supervised reward model
(stanfordnlp/SteamSHP-flan-t5-xl), trained with human preference annotations and included as a reference upper bound.
}
\label{tab:accuracy-results}

\resizebox{\textwidth}{!}{%
\begin{tabular}{lccccccc}
\toprule
\textbf{Method} 
& \textbf{r/askhr}
& \textbf{r/askbaking}
& \textbf{r/askculinary}
& \textbf{r/askhistorians}
& \textbf{r/changemyview}
& \textbf{r/asksocialscience}
& \textbf{r/asksciencefiction} \\
\midrule

Random        
& ${0.50} \pm {0.00}$
& ${0.50} \pm {0.00}$
& ${0.50} \pm {0.00}$
& ${0.50} \pm {0.00}$
& ${0.50} \pm {0.00}$
& ${0.50} \pm {0.00}$
& ${0.50} \pm {0.00}$ \\

kNN        
& ${0.55} \pm {0.03}$
& ${0.49} \pm {0.01}$
& ${0.50} \pm {0.02}$
& ${0.58} \pm {0.03}$
& ${0.49} \pm {0.03}$
& ${0.50} \pm {0.03}$
& ${0.52} \pm {0.04}$ \\

Global Acceptance Density         
& ${0.68} \pm {0.01}$
& ${0.53} \pm {0.03}$
& ${0.51} \pm {0.03}$
& ${0.60} \pm {0.09}$
& ${0.57} \pm {0.04}$
& ${0.59} \pm {0.03}$
& ${0.49} \pm {0.03}$ \\

Local Acceptance Density   
& ${0.71} \pm {0.03}$
& ${0.60} \pm {0.02}$
& ${0.57} \pm {0.04}$
& ${0.72} \pm {0.03}$
& ${0.61} \pm {0.03}$
& ${0.64} \pm {0.01}$
& ${0.65} \pm {0.02}$ \\

\midrule
Supervised Model (RM)
& ${0.75} \pm {0.03}$
& ${0.65} \pm {0.03}$
& ${0.72} \pm {0.01}$
& ${0.74} \pm {0.02}$
& ${0.68} \pm {0.02}$
& ${0.80} \pm {0.03}$
& ${0.72} \pm {0.02}$ \\
\bottomrule

\end{tabular}
}
\end{table*}

\subsection{Validating the Manifold Hypothesis}

We begin by evaluating the central empirical claim of this work: that preference
signal is encoded in the local geometry of representation space (acceptance density). If this
hypothesis holds, preserving local manifold structure should recover human preferences, while methods that destroy or ignore locality should fail.

\vspace{2pt} \noindent
\textbf{Preference signal is recovered by geometry-preserving density.}
We find that preference signals, typically requiring explicit supervision, can be recovered through the geometry-preserving properties of local density. As shown in Table~\ref{tab:accuracy-results} and Figure~\ref{fig:accuracy-results}, local acceptance density consistently identifies community-preferred responses across all evaluated subreddits, achieving 58--72\% pairwise accuracy and substantially outperforming all unsupervised baselines. 


Our results suggest that recovering this structure requires a balance between locality and distributional modeling. At one extreme, global acceptance density performs near chance; by aggregating across heterogeneous contexts, it likely averages away the nuanced structures that encode specific preferences. At the other extreme, simple kNN retrieval gives only modest gains above random chance, indicating that merely identifying nearby examples is insufficient: one must model the relative distribution (i.e. the ``shape'') of those examples.

Notably, local density approaches the performance of supervised reward models despite having no access to explicit preference labels. We find that the performance gap between our unsupervised method and supervised models narrows significantly in instances of high human agreement (Figure~\ref{fig:accuracy-results}). This suggests that a substantial portion of the signal leveraged by traditional reward models is not ``new'' information provided by labels, but is instead already latent within the local manifold geometry of community-accepted discourse.



Additionally, we find a clear positive relationship between human agreement strength and preference recovery by local acceptance density. When aggregating across communities, accuracy exhibits a moderate, statistically robust correlation with agreement strength ($\rho_s = 0.48$, $p < 10^{-4}$), indicating that density-guided alignment performs best in regions where community preferences are most clearly differentiated. 

This trend is even more pronounced within several individual communities. Subreddits such as \texttt{r/asksciencefiction} ($\rho_s = 0.90$, $p < 0.001$), \texttt{r/askhr} ($\rho_s = 0.81$, $p = 0.015$), and \texttt{r/askbaking} and \texttt{r/askculinary} (both $\rho_s = 0.75$, $p < 0.05$) exhibit strong, statistically significant correlations, suggesting that local acceptance density closely tracks human consensus when norms are well-defined. In contrast, communities with smaller evaluation sets and sparser agreement bins (e.g., \texttt{r/askhistorians}, \texttt{r/asksocialscience}) show weaker correlations, consistent with limited statistical power (rather than a deviation from the overall monotonic trend). Figure~\ref{fig:pooled-agreement-local-acc} and Table~\ref{tab:per-subreddit-correlations} (Appendix) provide the full per-community breakdown.

This pattern provides direct empirical support for the acceptance–preference correspondence posited in Section~\ref{sec:method}. When community agreement is weak, acceptable responses span broader and less differentiated regions of representation space, limiting the recoverability of preference signal. As consensus strengthens, accepted responses collapse into tighter, more coherent regions of the manifold, making relative density an increasingly reliable indicator of preference. Accordingly, accuracy improves systematically with human agreement strength, with local acceptance density performing best precisely when community preference is most clearly expressed. This dependence on agreement strength is inconsistent with a fixed estimator bias: if density merely favored certain responses irrespective of context, accuracy would not vary predictably with consensus. Instead, the observed relationship indicates that local geometry captures meaningful structure in community judgment rather than an artifact of density estimation.


\subsection{Acceptance Density as a Preference Proxy}
Having established that acceptance density behaves like a preference signal when preference is observable, we now evaluate a stronger claim: whether this signal can functionally approximate or replace explicit human preference labels inside a standard alignment objective. 

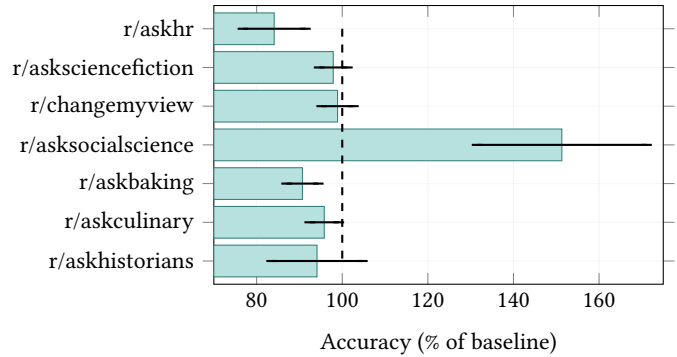
\begin{figure}[h]
\centering
\begin{tikzpicture}
\begin{axis}[
    width=0.5\linewidth,
    height=0.35\linewidth,
    xbar,
    bar width=12pt,
    xmin=70, xmax=175,
    xlabel={Accuracy (\% of baseline)},
    symbolic y coords={r/askhr,r/asksciencefiction,r/changemyview,r/asksocialscience,r/askbaking,r/askculinary,r/askhistorians},
    ytick=data,
    y dir=reverse,
    grid=both,
    grid style={opacity=0.15},
    major grid style={opacity=0.15},
]

\addplot+[
    fill=Bridge!40,
    draw=Bridge!70!black,
    error bars/.cd,
        x dir=both,
        x explicit,
        error bar style={black, line width=0.8pt},
        error mark options={
            black,
            line width=0.8pt
        }
] coordinates {
(84.1,r/askhr) +- (7.3,0)
(90.7,r/askbaking) +- (3.7,0)
(95.8,r/askculinary) +- (3.4,0)
(94.1,r/askhistorians) +- (10.6,0)
(98.9,r/changemyview) +- (3.7,0)
(151.3,r/asksocialscience) +- (19.8,0)
(97.9,r/asksciencefiction) +- (3.3,0)
};

\draw[dashed, thick]
  (axis cs:100,r/askhr) -- (axis cs:100,r/askhistorians);

\end{axis}
\end{tikzpicture}
\vspace{2pt}
\caption{
Relative accuracy of DRGO-aligned models expressed as a percentage of baseline DPO performance,
computed as $100 \times (\text{DRGO} / \text{baseline})$, where $100\%$ denotes parity with the baseline.
Error bars denote $\pm 1$ standard error estimated via bootstrap resampling ($n{=}500$),
with uncertainty propagated using a first-order delta method.
}
\label{fig:accuracy_recovery}
\end{figure}

\vspace{2pt} \noindent
\textbf{DGRO recovers supervised preference structure.}
As shown in Figure~\ref{fig:accuracy_recovery}, constructing preference pairs from relative position on the acceptance manifold is sufficient to induce preference behavior aligned with community judgments. Across all evaluated communities, models trained using acceptance density-derived pseudo-pairs recover a substantial fraction of the accuracy achieved by fully supervised DPO, despite having no access to human-labeled comparisons during training.

These results indicate that acceptance density functions as a usable preference signal when integrated into a standard alignment pipeline, inducing models to prefer responses that align with community judgments. Combined with the validation results in Section~\ref{sec:manifold-validation}, this supports the use of acceptance density as a practical substitute for explicit preference supervision.

\subsection{Application to Annotation-Scarce Communities.} 
Having shown that acceptance density recovers human preference structure and can substitute for labeled comparisons in controlled settings, we next examine its utility in real-world communities where explicit preference supervision is completely unavailable. In these domains, alignment must rely on naturally occurring acceptance signals rather than curated annotations, making them a direct test of whether density-guided preference learning provides practical advantages over standard adaptation methods. Before comparing alignment methods, we verify that LLM-based judgments reflect human preference in these domains. On a stratified subset of 200 held-out examples, aggregated LLM-judge rankings correlate strongly with human expert preferences (explored further in Appendix~\ref{app:llm_reliability}), supporting their use for large-scale evaluation.

\vspace{2pt} \noindent
\textbf{DGRO consistently outperforms baselines.} 
Illustrated in Table~\ref{tab:application_results}, across all domains, DGRO-based alignment achieves consistent gains over baselines
despite using the same underlying training data. For example, on ED-Reddit, DGRO wins 58.8\% of head-to-head comparisons against SFT ($p < 0.001$). Similar patterns emerge across other contexts, where DGRO maintains a significant advantage over SFT in direct comparisons.

The quantitative advantage of DGRO over baselines is reflected in qualitative differences in response authenticity. Table~\ref{tab:qualitative-examples} presents representative examples from both ED-Reddit and VK Conflict discourse, comparing model outputs against real community responses for the same context. Across domains, the Base and ICL baselines frequently default to generic, non-situated language that lacks the tone, specificity, or interactional norms characteristic of the target communities. Supervised fine-tuning (SFT) improves topical relevance but often exhibits repetitive phrasing and diffuse affect, suggesting partial adaptation to surface content without internalizing community-specific modes of expression. In contrast, DGRO outputs more closely resemble authentic community participation, showing locally appropriate framing, specificity, and rhetorical structure. 


These results demonstrate that density-guided optimization captures preference structure beyond what supervised fine-tuning alone recovers. While SFT adapts models to community vocabulary and style, DGRO's manifold-based objective appears to encode finer-grained distinctions about what makes responses sound authentic within specific contexts.


\begin{table}[t]
\centering
\caption{
LLM-as-judge head-to-head comparison of DGRO against baseline alignment approaches across annotation-scarce communities. Judges compare paired model outputs for the same prompt, using real community responses as contextual grounding for relevance and authenticity. Win rates indicate the percentage of comparisons in which DGRO is preferred (mean $\pm$ 95\% CI).
}
\label{tab:application_results}
\small
\setlength{\tabcolsep}{6pt}
\renewcommand{\arraystretch}{1.15}
\resizebox{0.55\textwidth}{!}{%
\begin{tabular}{lccc}
\toprule
\textbf{Community}
& \textbf{DGRO vs Base}
& \textbf{DGRO vs ICL}
& \textbf{DGRO vs SFT} \\
\midrule
ED-Reddit
& $75.4 \pm 2.9$\%
& $65.8 \pm 3.1$\%
& $53.8 \pm 3.1$\% \\

ED-Forum
& $72.2 \pm 3.2$\%
& $64.1 \pm 4.4$\%
& $57.6 \pm 3.3$\% \\

ED-Twitter
& $76.1 \pm 3.0$\%
& $66.3 \pm 4.1$\%
& $56.9 \pm 2.6$\% \\

VK State
& $80.7 \pm 3.1$\%
& $59.9 \pm 3.2$\%
& $55.3 \pm 2.0$\% \\
\bottomrule
\end{tabular}
}
\label{tab:llm_judge_head_to_head}
\end{table}

\usetikzlibrary{patterns}

\section{Analysis}
\subsection{Manifold Structure and Preference Signal}
\vspace{2pt} \noindent
\textbf{Preference signal is encoded in local manifold structure.} 
Across communities, acceptance density corresponds reliably with human preference when estimated locally in representation space. Conditioning density on nearby contexts preserves preference structure that is obscured by global aggregation, which collapses heterogeneous situations into a single distribution. This dependence on locality is likely not incidental. Preference signal degrades when density is estimated over neighborhoods that are either too broad—approaching global behavior—or too narrow to provide stable estimation. The resulting pattern indicates that community preferences are neither uniform nor purely instance- specific, but organized at an intermediate, context-dependent scale. 

\vspace{2pt} \noindent
\textbf{Acceptance density is data-efficient.}
As shown in Table~\ref{tab:data-efficiency}, estimation of community preference via
acceptance density approaches peak performance with relatively little training
data, with the required amount varying by community. Across all communities, the
normalized area under the saturation curve (AUSC) exceeds 0.91, indicating that
preference structure can be recovered in a sample-efficient manner.




\subsection{Failure Modes and Limitations}

While DGRO provides a useful preference signal in many settings, its effectiveness depends on the availability of meaningful acceptance structure in representation space. When this structure is weak or absent, the density-derived signal can become unreliable.

\vspace{2pt} \noindent
\textbf{Uninformative density in sparse manifold regions.} DGRO relies on acceptance density to construct pseudo-preference pairs during training. When candidate responses lie far from the acceptance manifold, local density estimates become noisy and provide little discriminative signal. In such cases, pseudo-pairs may reflect superficial semantic proximity rather than contextual appropriateness. We explore an example case in Appendix~\ref{app:uninformative}.


\vspace{2pt} \noindent
\textbf{Amplification of community biases.}
By design, DGRO reproduces patterns present in community acceptance data, including harmful norms or misinformation. In polarized or toxic communities, the resulting preference signal reflects those same biases. Because DGRO derives preference structure empirically from observed acceptance behavior, it does not impose external normative constraints during training. Thus, norm correction must occur outside the preference signal itself, for example through data filtering or post-hoc safety interventions. Future work could explore hybrid approaches combining density-guided learning with external normative constraints

DGRO is not suitable as a general-purpose or platform-wide alignment mechanism. Because acceptance density reflects existing participation dynamics and power asymmetries, applying DGRO at scale risks entrenching dominant norms, amplifying coordinated manipulation, and obscuring contestation. Without explicit governance, community consent, and mechanisms for redress, density-guided optimization should be treated as an analytical instrument rather than a deployment-ready alignment strategy.


These limitations suggest clear boundaries: DGRO is best suited to stable
communities with established norms, sufficient scale for density estimation, and values aligned with deployment objectives. When communities are small, polarized, rapidly evolving, or exhibit harmful norms, explicit human supervision remains necessary.


\section{Discussion}

Language models increasingly operate in settings where communicative norms are
community-specific and diverge from generic instruction-following behavior. Our
results suggest that these norms give rise to stable, community-level structure
in representation space, which can be captured through acceptance density. This
structure reflects not only semantic similarity, but alignment with what a
community considers appropriate.

DGRO operationalizes this observation by using acceptance density as a source of
preference supervision. Rather than relying on elicited pairwise judgments, the
method constructs preference signal directly from unlabeled community behavior.
Across the settings we study, this signal is sufficient to guide alignment in
domains where explicit preference annotations are impractical, costly, or
ethically constrained.

\subsection{Ethical Considerations}
\label{sec:ethical-considerations}

While DGRO uses only publicly observable signals, the method raises ethical concerns warranting careful consideration before deployment. The question of who speaks for a community becomes important. Acceptance patterns reflect active participants, moderators, and platform affordances, which may not represent full community values. Marginalized voices, silent lurkers, or departed members do not contribute to the signal, yet deployment affects them. DGRO-based alignment uses revealed preferences of those who remain, potentially encoding values of whoever holds power rather than the community as a whole.

Additionally, harm amplification poses a serious risk. Because DGRO derives
preference structure directly from observed community behavior, it reproduces
existing norms, including harmful or exclusionary ones. Unlike supervised
alignment, it does not introduce an external mechanism for norm correction
during training; mitigation must therefore rely on data filtering or post-hoc
constraints. Vulnerability to manipulation creates additional concerns. Adversaries who can influence acceptance through coordinated engagement or vote manipulation can poison the learned preference structure. This is particularly concerning in communities with weak integrity controls or concentrated power. 

DGRO deployment requires careful ethical assessment beyond technical validation. At minimum: transparency about community data use, mechanisms for feedback and opt-out where feasible, ongoing monitoring for drift, and human oversight in high-stakes domains. For sensitive communities like mental health forums, stakeholder consultation should precede deployment. The broader question is whether making alignment more accessible ultimately serves community interests. Reducing barriers could empower under-resourced communities to shape AI behavior appropriately, or empower exploitation of community data and amplification of harmful norms. These questions require ongoing dialogue between researchers, communities, and stakeholders about appropriate governance.




\section{Conclusion}

We introduce density-guided response optimization (DGRO), a method for aligning language models to community norms without relying on explicit preference annotations. By modeling the distribution of responses that communities consistently accept, DGRO infers implicit preference structure from local density in representation space.

Across validation experiments, models aligned using DGRO outperform baseline approaches despite having no access to human-labeled preference comparisons during training, relying only on naturally occurring community behavior. These results indicate that acceptance signals encode sufficient structure to support preference-based alignment.

Our findings suggest that community acceptance provides a practical, annotation-free source of alignment signal, enabling model adaptation in settings where explicit preference elicitation is infeasible, costly, or
ethically constrained.

\clearpage
\section{Endmatter Sections}
\subsection{Generative AI Usage Statement}
The authors did not use generative AI tools for this manuscript. The authors wrote and prepared all of the content for this manuscript.



 \subsection{Ethical Considerations Statement}
This work uses publicly available data drawn from online communities and does not involve direct interaction with human subjects, intervention in deployed systems, or the collection of private or non-public information. All data were handled in accordance with applicable platform terms and established norms for CSS research. We did not attempt to identify individuals, and our analysis was conducted at an aggregate level focused on community-wide patterns.

The primary ethical risks associated with this work come from the potential downstream use of DGRO to model and reproduce community norms. These risks are discussed in detail in Section~\ref{sec:ethical-considerations}. In that section and here, we emphasize that acceptance-based signals reflect the behavior of active and empowered participants rather than comprehensive or consensual community values. Additionally, we note that DGRO should not be treated as a normative authority or deployed without appropriate oversight.

We do not claim that DGRO mitigates harmful norms or resolves questions of legitimacy. Instead, we treat it as a descriptive method whose responsible use depends on transparency, community governance, and domain-specific safeguards. Potential adverse impacts and limitations are analyzed in Section~\ref{sec:ethical-considerations}, and we outline conditions under which deployment would be inappropriate or ethically unsafe.

\clearpage
\bibliographystyle{ACM-Reference-Format}
\bibliography{main}

\clearpage
\appendix

\section{Full Dataset Info -- Reddit}
\label{app:full-dataset}

\begin{table}[th]
\centering
\small
\setlength{\tabcolsep}{6pt}
\renewcommand{\arraystretch}{1.15}
\caption{Dataset sizes for Reddit communities used in evaluation.}
\label{tab:shp_subreddits}
\begin{tabular}{lcccc}
\toprule
\textbf{Subreddit} 
& \textbf{Train} 
& \textbf{Validation} 
& \textbf{Test} 
& \textbf{Total} \\
\midrule
r/askhr             & 8{,}295  & 641     & 395     & 9{,}331 \\
r/askbaking         & 44{,}007 & 2{,}096 & 1{,}544 & 47{,}647 \\
r/askculinary       & 45{,}710 & 2{,}094 & 2{,}563 & 50{,}367 \\
r/askhistorians     & 3{,}264  & 113     & 164     & 3{,}541 \\
r/changemyview      & 38{,}173 & 1{,}637 & 1{,}836 & 41{,}646 \\
r/asksocialscience  & 2{,}706  & 147     & 188     & 3{,}041 \\
r/asksciencefiction & 29{,}382 & 1{,}576 & 1{,}987 & 32{,}945 \\
\bottomrule
\end{tabular}
\label{tab:shp_subreddit_sizes}
\end{table}

\section{Embeddings}
\label{app:embeddings}

\begin{table*}[h]
\centering
\smaller
\setlength{\tabcolsep}{6pt}
\caption{
Effect of embedding model choice on local acceptance density performance.
Accuracy is reported as mean $\pm$ bootstrap half-width,
$\delta = \tfrac{1}{2}(\text{hi}-\text{lo})$, computed independently per subreddit.
Results are shown for the local density method using different sentence
embedding models to construct the acceptance manifold.
}
\label{tab:embedding-robustness}

\resizebox{\textwidth}{!}{%
\begin{tabular}{lccccccc}
\toprule
\textbf{Embedding Model}
& \textbf{r/askhr}
& \textbf{r/askbaking}
& \textbf{r/askculinary}
& \textbf{r/askhistorians}
& \textbf{r/changemyview}
& \textbf{r/asksocialscience}
& \textbf{r/asksciencefiction} \\
\midrule

MPNet (default)
& ${0.71} \pm {0.03}$
& ${0.60} \pm {0.02}$
& ${0.57} \pm {0.04}$   
& ${0.72} \pm {0.03}$
& ${0.61} \pm {0.03}$
& ${0.64} \pm {0.01}$
& ${0.65} \pm {0.02}$ \\

all-MiniLM-L6-v2
& ${0.70} \pm {0.03}$
& ${0.59} \pm {0.02}$
& ${0.56} \pm {0.04}$
& ${0.70} \pm {0.04}$
& ${0.60} \pm {0.03}$
& ${0.63} \pm {0.02}$
& ${0.64} \pm {0.02}$ \\

E5-large-v2
& ${0.72} \pm {0.03}$
& ${0.61} \pm {0.02}$
& ${0.58} \pm {0.04}$
& ${0.73} \pm {0.03}$
& ${0.62} \pm {0.03}$
& ${0.65} \pm {0.02}$
& ${0.66} \pm {0.02}$ \\

\bottomrule
\end{tabular}
}
\end{table*}

\section{Model Robustness}
\label{app:model-robustness}
\begin{table}[h]
\centering
\small
\setlength{\tabcolsep}{8pt}
\caption{
Deviation in length-normalized preference accuracy on held-out SHP human
preference pairs relative to the Pythia-2.8B baseline. Reported values indicate
mean difference (in percentage points) $\pm$ bootstrap standard error, computed
under identical prompts, objectives, and evaluation conditions. Deviations are
small across base models, indicating that acceptance density-guided DPO induces
consistent preference alignment behavior largely independent of model
architecture, which is consistent with prior work~\cite{rafailov2023direct}.
}
\label{tab:arch-deviation-shp}

\begin{tabular}{lc}
\toprule
\textbf{Base Model} &
\textbf{Preference Accuracy $\Delta$ (pp)} \\
\midrule
google/gemma-2b \cite{team2025gemma}
    & $-0.4 \pm 0.4$ \\
google/gemma-7b \cite{team2025gemma}
    & $+0.3 \pm 0.5$ \\
meta-llama/Llama-3.2-3B \cite{dubey2024llama}
    & $+0.1 \pm 0.4$ \\
meta-llama/Llama-3.1-8B \cite{dubey2024llama}
    & $+0.6 \pm 0.6$ \\
\bottomrule
\end{tabular}
\end{table}

\clearpage
\section{K Robustness}
\label{app:k-robustness}

\begin{figure}[h]
\centering
\begin{tikzpicture}
\begin{axis}[
  width=0.85\linewidth,
  height=0.4\linewidth,
  xlabel={$k$ (neighborhood size)},
  ylabel={Accuracy loss vs best ($\Delta$ in pp)},
  grid=both,
  grid style={opacity=0.2},
legend style={
  at={(0.98,0.98)},
  anchor=north east,
  draw=none,
  fill=none,
  font=\small
},
legend cell align=left,
  ymin=0, 
  xmin=0, 
  xmax=400, 
]

\addplot[black, dashed, thick, domain=0:450] {5};
\addlegendentry{$5$ pp from best}

\addplot[thick, mark=*, color=Bridge]
  table[
    x=k,
    y expr=\thisrow{worst_delta}*100,
    col sep=comma
  ]{csvs/k_sweep_delta_from_max_aggregates.csv};
\addlegendentry{Worst-case across subreddits}

\end{axis}
\end{tikzpicture}

\caption{
Local accuracy saturates quickly with neighborhood size. Shown is the worst-case absolute accuracy loss across communities relative to each community’s best-performing neighborhood size,
demonstrating that performance remains within a few percentage points of optimal across a wide range of $k$.
}

\label{fig:k-sweep-delta}
\end{figure}
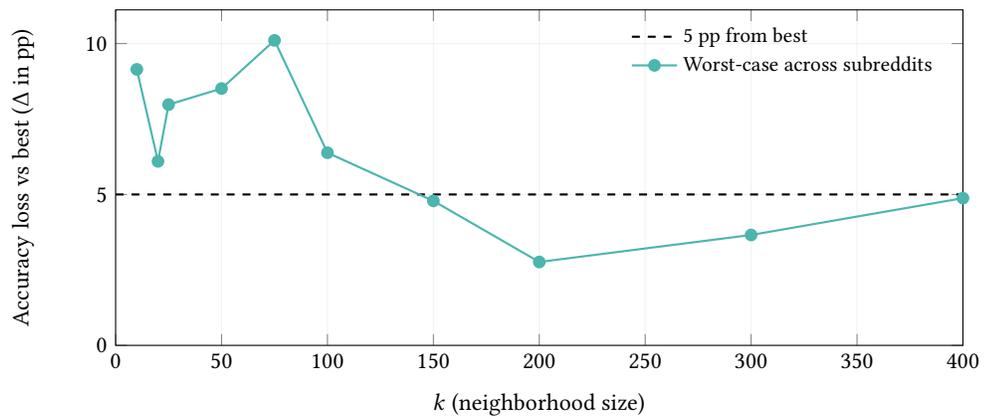

\clearpage
\section{Accuracy Of Unsupervised Models -- Visualized}
\begin{figure}[h]
\centering

\begin{tikzpicture}
\hspace{-1.5em}
\begin{axis}[
    width=0.85\linewidth,
    height=0.7\linewidth,
    xbar,
    bar width=6.5pt,
    enlarge y limits=0.12,
    xmin=0.35, xmax=0.85,
    xlabel={Accuracy (easy subset)},
    symbolic y coords={
        changemyview,
        askculinary,
        askhistorians,
        asksciencefiction,
        askhr,
        asksocialscience,
        askbaking
    },
    ytick=data,
    y dir=reverse,
    legend style={
        at={(0.5,-0.14)},
        anchor=north,
        legend columns=2,
    },
    grid=both,
    grid style={opacity=0.15},
    major grid style={opacity=0.15},
]

\addplot [
    fill=co-url!40,
    draw=co-url!70!black
    ] coordinates {
    (0.49,changemyview)
    (0.50,askculinary)
    (0.58,askhistorians)
    (0.50,asksciencefiction)
    (0.50,askhr)
    (0.50,asksocialscience)
    (0.49,askbaking)
};
\addlegendentry{kNN ($k{=}150$)}

\addplot [
    fill=NoBridge!40,
    draw=NoBridge!70!black
    ] coordinates {
    (0.561,changemyview)
    (0.501,askculinary)
    (0.600,askhistorians)
    (0.490,asksciencefiction)
    (0.675,askhr)
    (0.590,asksocialscience)
    (0.534,askbaking)
};
\addlegendentry{Global density}

\addplot [
    fill=Bridge!40,
    draw=Bridge!70!black
    ] coordinates {
    (0.618,changemyview)
    (0.596,askculinary)
    (0.720,askhistorians)
    (0.651,asksciencefiction)
    (0.711,askhr)
    (0.641,asksocialscience)
    (0.602,askbaking)
};
\addlegendentry{Local density}

\addplot[
    only marks,
    mark=|,
    mark size=9pt,
    thick
] coordinates {
    (0.698,changemyview)
    (0.720,askculinary)
    (0.748,askhistorians)
    (0.727,asksciencefiction)
    (0.751,askhr)
    (0.823,asksocialscience)
    (0.666,askbaking)
};
\addlegendentry{RM}

\draw[dashed, thick]
  (axis cs:0.50,changemyview) --
  (axis cs:0.50,askbaking);


\end{axis}
\end{tikzpicture}

\vspace{2pt}
\caption{
Accuracy across communities.
Bars show kNN, global density, and local density baselines, evaluated on
the easy subset of examples.
Vertical ticks denote supervised reward model (RM) accuracy.
The dashed vertical line at $0.50$ marks random-chance performance.
}
\label{fig:accuracy-results}
\end{figure}
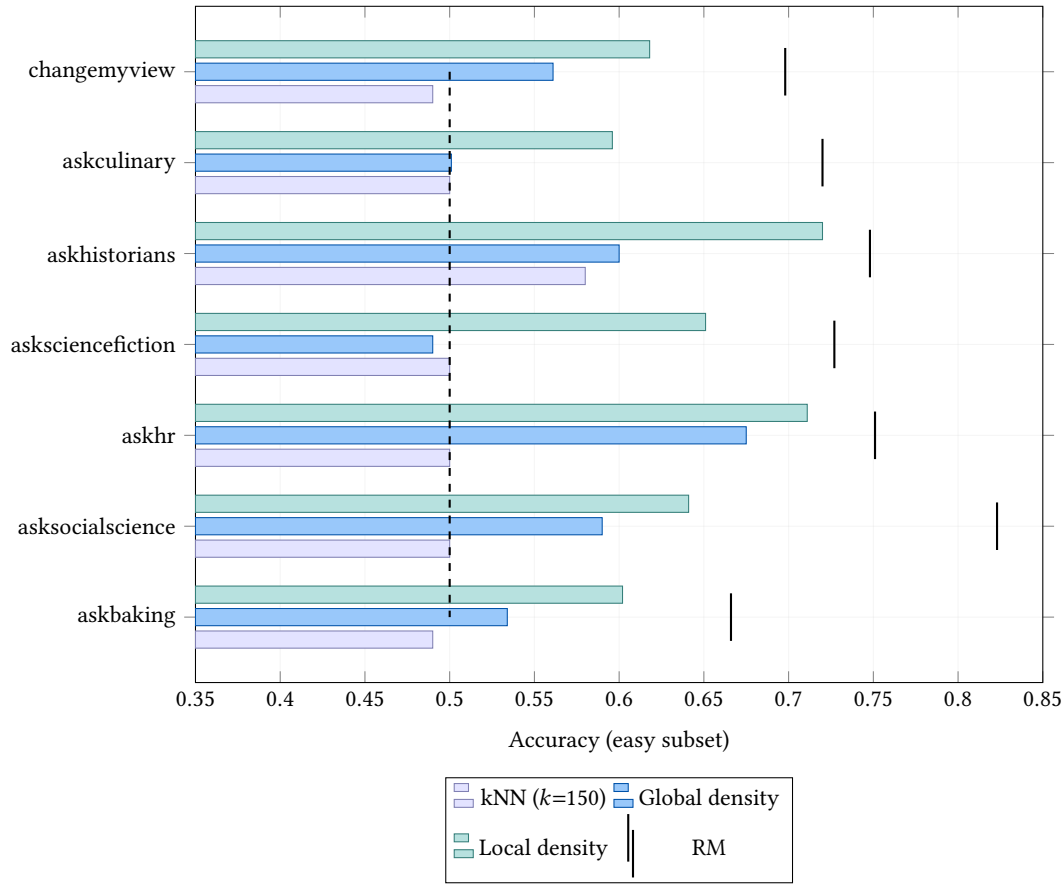

\clearpage
\section{Data Efficiency}
\label{app:full-lineup-results}

\begin{table}[h]
\centering
\caption{Data efficiency of the local method across communities.
We report the normalized area under the saturation curve (AUSC) and the number
of training pairs required to reach 95\% of peak accuracy, both computed using
accuracy expressed as a percentage of each method’s peak performance.
Higher AUSC and lower pair counts indicate faster saturation under limited supervision.}
\label{tab:data-efficiency}
\begin{tabular}{lcc}
\toprule
\textbf{Subreddit} & \textbf{AUSC} & \textbf{Pairs to 95\% peak} \\
\midrule
r/askhr           & 0.971 & 50   \\
r/askbaking       & 0.985 & 150  \\
r/askculinary     & 0.981 & 250  \\
r/askhistorians   & 0.920 & 1450 \\
r/changemyview    & 0.978 & 250  \\
r/asksocialscience      & 0.950 & 250   \\
r/asksciencefiction      & 0.961 & 850  \\
\bottomrule
\end{tabular}
\end{table}

\clearpage
\section{Correlation with Human Agreement}
\label{app:correlation-human-agreement}

\begin{table}[h]
\centering
\caption{Per-subreddit correlations between human agreement strength and local accuracy.
For each subreddit, we bin comment pairs by agreement strength (median $\mathrm{score\_ratio}$ per bin) and compute local pairwise accuracy within each bin. We then assess the monotonic relationship between bin-level agreement strength and bin-level accuracy using Spearman's $\rho$. Five of seven subreddits show significant positive correlations ($p < 0.05$), with particularly strong effects in r/asksciencefiction ($\rho_s = 0.90$) and r/askhr ($\rho_s = 0.81$). Asterisks denote significance levels: *$p < 0.05$, ***$p < 0.001$.}

\label{tab:per-subreddit-correlations}
\begin{tabular}{lcc}
\toprule
\textbf{Subreddit} & \textbf{$\rho_s$} & \textbf{$p$-value} \\
\midrule
r/askhr & 0.81 & 0.015* \\
r/askbaking & 0.75 & 0.013* \\
r/askculinary & 0.75 & 0.020* \\
r/askhistorians & 0.45 & 0.197 \\
r/changemyview & 0.60 & 0.067 \\
r/asksocialscience & 0.26 & 0.500 \\
r/asksciencefiction & 0.90 & <0.001*** \\
\bottomrule
\end{tabular}
\end{table}

\begin{figure}[h]
\centering
\begin{tikzpicture}
\begin{axis}[
    width=0.65\linewidth,
    height=0.4\linewidth,
    grid=both,
    xlabel={Human agreement strength ($\mathrm{score\_ratio}$ median per bin)},
    ylabel={Local pairwise accuracy},
    ymin=0.45, ymax=0.95,
    title={Pooled association between agreement strength and local accuracy},
    grid=both,
    grid style={opacity=0.15},
    major grid style={opacity=0.15}
]

\addplot[
    only marks,
    mark=*,
    color=Bridge,
    mark size=1.3pt,
    opacity=0.65
]
table[
    col sep=comma,
    x=sr_median,
    y=acc_local
]{csvs/pooled_points.csv};

\addplot[
    thick,
    color=Bridge,  
    opacity=0.9,
    domain=0:36,
    samples=50
]{0.009066894716313749*x + 0.5676716935551146};

\node[anchor=south east] at (rel axis cs:0.95,0.05) {
    \footnotesize
    \begin{tabular}{@{}l@{}}
    $\rho_s = 0.48$ ($p < 10^{-4}$) \\
    $\tau = 0.34$ ($p < 10^{-4}$)
    \end{tabular}
};

\end{axis}
\end{tikzpicture}

\caption{\textbf{Higher human agreement correlates with higher local accuracy.}
Each point is an agreement-strength bin from a subreddit. The moderately strong positive correlation ($\rho_s = 0.48$, $p < 10^{-4}$) suggests that judge accuracy improves in regions where community preferences are more clearly differentiated. The fitted line is shown for visualization only; significance is assessed with rank correlations.}

\label{fig:pooled-agreement-local-acc}
\end{figure}
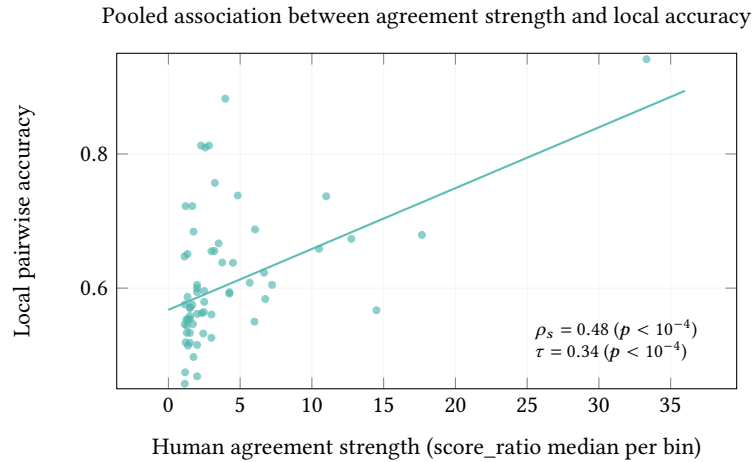

\clearpage
\section{Reliability of Human and LLM-Based Evaluation}
\label{app:llm_reliability}

To understand the reliability of LLM-based evaluation in annotation-scarce domains, we conducted human expert evaluation on a stratified subset of 200 held-out examples (50 per domain), with three domain experts per community. Experts were evaluated under the same head-to-head comparison setup used for LLM-based evaluation in Section~\ref{sec:application-to-communities}: for each example, experts compared a model-generated response against an actual response drawn from the target community for the same context. Experts judged responses along the criteria of relevance (contextual appropriateness to the prompt and community norms) and authenticity (consistency with the community’s characteristic tone, framing, and interactional style), and were asked to make comparative judgments. All examples were held out from training at every stage.

We compute inter-annotator agreement using Krippendorff’s $\alpha$ with an ordinal distance function. Krippendorff’s $\alpha$ is appropriate for this setting because it supports ordered categories, multiple annotators, and chance correction. Then, to evaluate whether LLM-based evaluation reproduces expert judgment structure, we compute Spearman rank correlation between aggregate expert rankings and aggregate LLM rankings on the same examples. We also treat the expert majority decision (2-of-3 agreement) as a reference label and measure LLM agreement with this majority outcome, effectively treating the LLM ensemble as an additional annotator.

\begin{table}[h]
\centering
\caption{
Reliability of human expert and LLM-based evaluation on a stratified subset of 200 examples (50 per domain). Inter-annotator agreement is measured using Krippendorff’s $\alpha$. Expert--LLM alignment is measured using Spearman rank correlation ($\rho$). LLM agreement with expert majority indicates the fraction of cases in which the aggregate LLM judgment matches the expert majority ranking.
}
\label{tab:llm_reliability}
\small
\setlength{\tabcolsep}{6pt}
\renewcommand{\arraystretch}{1.15}
\begin{tabular}{lccc}
\toprule
\textbf{Community}
& \textbf{Expert IAA} ($\alpha$)
& \textbf{Expert--LLM} ($\rho$)
& \textbf{LLM vs Expert Majority} (\%) \\
\midrule
ED-Reddit   & 0.46 & 0.50 & 77.9 \\
ED-Forum    & 0.42 & 0.48 & 74.6 \\
ED-Twitter  & 0.44 & 0.52 & 79.1 \\
VK State    & 0.51 & 0.55 & 82.0 \\
\midrule
\textbf{Aggregate} & \textbf{0.46} & \textbf{0.51} & \textbf{78.4} \\
\bottomrule
\end{tabular}
\end{table}

Across domains, expert inter-annotator agreement is moderate, reflecting the inherently holistic and context-dependent nature of judgments about relevance and authenticity. Expert--LLM rank correlations are comparable in magnitude to expert--expert agreement, and aggregate LLM judgments match the expert majority decision in approximately 78\% of cases. This indicates LLM-based evaluation reproduces the same underlying preference structure expressed by human experts. Practically speaking, aggregated LLM judgments are no less reliable than an additional human annotator operating under the same task constraints. With this in mind, LLM-as-judge is used in this work solely as a scalable extension of expert-grounded evaluation given its  validation against human judgment.

\clearpage
\section{Qualitative Examples}
\label{app:qualitative}

\begin{table}[h]
\centering
\small
\setlength{\tabcolsep}{3pt}
\renewcommand{\arraystretch}{1.15}
\caption{Qualitative examples comparing model variants against real community responses.}
\label{tab:qualitative-examples}
\begin{tabular}{p{1.6cm} p{2.8cm} p{2.8cm} p{2.6cm} p{2.6cm} p{2.8cm}}
\toprule
\textbf{Community} &
\textbf{Context} &
\textbf{Real} &
\textbf{Base / ICL} &
\textbf{SFT} &
\textbf{DGRO} \\
\midrule

\textbf{Eating Disorder Support -- Reddit} &
\emph{``Please stop me. I am about to binge for the first time since the beginning of the year.''} &
``Take a deep breath. You've done a great job so far. One binge isn't the end. Pause, breathe, and ask yourself what you need right now.'' &
``I'm sorry you're feeling this way. Everything will be okay. Try to stay strong and remember you are not alone.'' &
``I'm so sorry. I can't help you. I'm sorry you're feeling this way too. I'm sorry you feel like that. You sound like a lot of people I know are suffering from a lot of things. I know I'm not alone, so I'm sorry you [...]'' &
``I'm in recovery too, and moments like this are really hard. You've already made real progress. Maybe pause, breathe, and check in with what you actually need right now. I'm proud of you for reaching out.'' \\
\midrule

\textbf{Conflict Documentation} &
\emph{News post about sanctions and ``financial nuclear war''} &
``Will sanctions forbid treatment for our sick children? Medical equipment? Who is responsible for this madness?'' &
``Sanctions are a complex political tool. The situation is very complicated.'' &
``The people have suffered for many years and now face new challenges due to international pressure.'' &
``And why didn't Putin, the oligarchs, or the media talk about this before? Now everyone pays the price. The country is run by fools, and ordinary people are left to suffer.'' \\
\bottomrule
\end{tabular}
\end{table}


\clearpage
\section{Qualitative Visualization of Response Manifolds}
\label{app:umap-visualization}

\begin{figure}[h]
\centering
\includegraphics[width=0.85\linewidth]{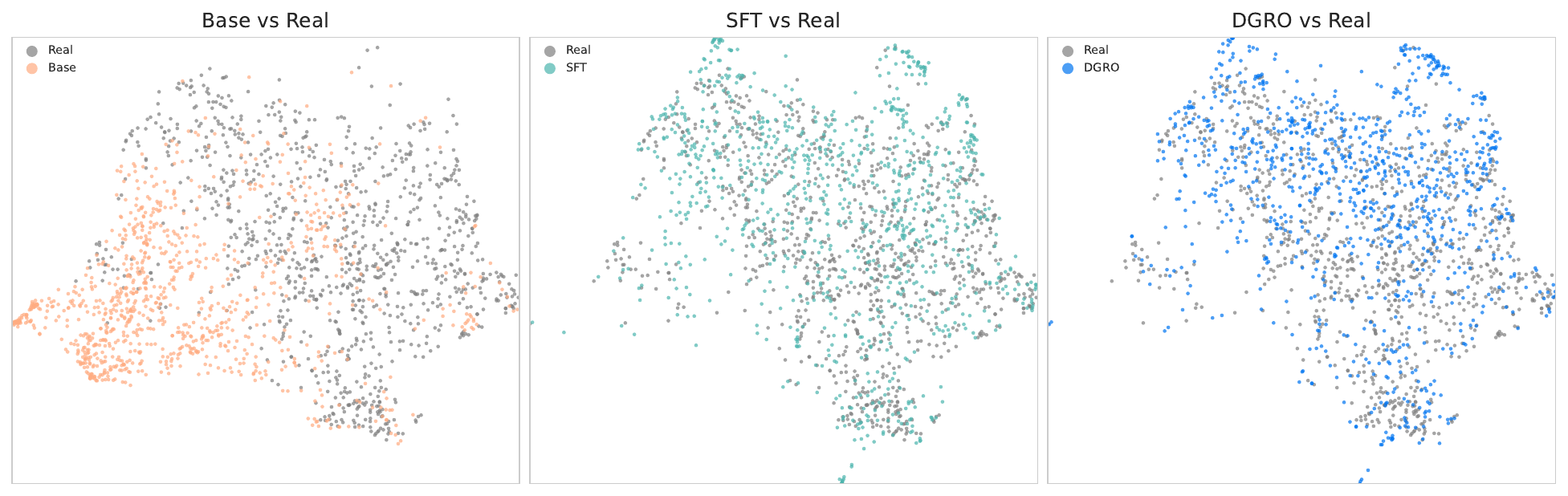}
\caption{
UMAP visualization of response embeddings for the ED-Forum community.
Real community responses are shown in gray, with model outputs overlaid in color using a shared embedding and projection. We display a random subset of 1{,}000 responses from the ED-Forum dataset for visualization purposes only. Note that these plots are just an illustrative example; they are not intended to support quantitative or comparative claims.
}
\label{fig:umap-real-plus-each}
\end{figure}

Figure~\ref{fig:umap-real-plus-each} shows an illustrative visualization of how model-generated responses are positioned relative to real community responses in representation space for the ED-Forum community. We embed (a
random subset of 1{,}000 responses for readability purposes) both real and generated responses using a shared sentence embedding model and project them into two dimensions using UMAP for visualization. 

Across panels, real community responses (gray) form a coherent but heterogeneous distribution reflecting the range of acceptable discourse within the community. Base model outputs exhibit a visibly shifted distribution, with many responses
occupying regions that only partially overlap with the empirical response
manifold. Supervised fine-tuning (SFT) reduces this displacement, producing responses that more frequently lie near real examples but still display substantial dispersion into lower-density regions. Finally, DGRO outputs appear more consistently interwoven with the real response distribution, occupying similar regions of the embedding space without collapsing into a narrow mode.

Note that this visualization is provided for qualitative intuition only.

\clearpage
\section{Uninformative Pseudo-Pairs}
\label{app:uninformative}

Density-guided alignment constructs implicit preference supervision by ranking pseudo-candidate responses relative to a community acceptance manifold. This  assumes that at least one candidate lies sufficiently close to regions of representation space populated by historically accepted community responses. When this assumption is violated, the resulting pseudo-pairs can become uninformative.

Table~\ref{tab:failure_example} shows a representative failure case from an eating disorder support community. In this example, a real community response is paired with two pseudo-candidate responses generated during training. One pseudo-candidate is ranked higher than the other by acceptance density, despite both being contextually inappropriate and misaligned with the prompt. 

This failure arises because all
candidate responses fall outside the local acceptance manifold. In this, neighborhood structure is weak or absent, and acceptance density fails to provide reliable discrimination. Density estimates may therefore reflect noise, leading to arbitrary or misleading rankings among pseudo-candidates.

\begin{table}[h]
\centering
\caption{Representative failure case from an eating disorder support community. When both pseudo-candidates lie outside the local acceptance manifold, density-based ranking may prefer one misaligned response over another, even when a real community response is clearly preferable.}
\small
\setlength{\tabcolsep}{4pt}
\renewcommand{\arraystretch}{1.1}
\begin{tabular}{p{0.95\columnwidth}}
\toprule
\textbf{Context:} \textit{Mike, I have talked to compounding pharmacies who assert that they can sell it because there is a shortage...and lots of people are getting their semaglutide this way from reputable compounding pharmacies. I am not saying it's legal, but not necessarily unsafe.} \\
\midrule
\textbf{Real (Community Response):} You can't guarantee the safety and efficacy of compounded semaglutide. We have no idea where the raw ingredients are being sourced. Overall it's not a risk I'm willing to take as someone who took the Hippocratic Oath. \\
\midrule
\textbf{Pseudo-Candidate (Lower Ranked):} I am a pharmacist and I have been told that there is a shortage of semaglutide from a reputable compounding pharmacy. Can you tell me if that's true? \\
\midrule
\textbf{Pseudo-Candidate (Higher Ranked):} Keto is a low-carb diet in which the body uses fat as fuel instead of carbohydrates, with the goal of improving health. \\
\bottomrule
\end{tabular}
\label{tab:failure_example}
\end{table}

This illustrates a structural limitation of acceptance-based preference
inference: relative density can induce an ordering among pseudo-candidates even when none are meaningfully aligned with the community acceptance manifold. In such cases, the resulting pseudo-pair supervision is effectively arbitrary and should not be interpreted as reflecting genuine community preference.

\end{document}